\documentclass[a4paper,10pt]{article}
\usepackage{amsmath,amssymb,natbib,mathtools}
\usepackage{newtxtext,newtxmath}
\usepackage[textwidth=5.5in,textheight=9in]{geometry}
\usepackage{hyperref}
\usepackage{url}
\usepackage{tikz}
\usepackage{authblk}
\usepackage{booktabs,multirow,colortbl}
\newcommand{\bftab}{\fontseries{b}\selectfont}
\usepackage{subcaption}

\title{Pushing Toward the Simplex Vertices: A Simple Remedy for Code Collapse in Smoothed Vector Quantization}
\author[1,2]{Takashi Morita\thanks{tmorita@alum.mit.edu}}
\affil[1]{Academy of Emerging Sciences, Chubu University}
\affil[2]{Institute for Advanced Research, Nagoya University}
\begin{document}
\maketitle

\begin{abstract}
  Vector quantization, which discretizes a continuous vector space into a finite set of representative vectors (a \emph{codebook}), has been widely adopted in modern machine learning.
Despite its effectiveness, vector quantization poses a fundamental challenge:
the non-differentiable quantization step blocks gradient backpropagation.
\emph{Smoothed} vector quantization addresses this issue by relaxing the discrete selection of a codebook vector into a weighted combination of codebook entries, represented as the matrix product of a simplex vector and the codebook.
Effective smoothing requires two properties:
(1) smoothed code-selection vectors should remain close to a onehot vector, ensuring tight approximation, and
(2) all codebook entries should be utilized, preventing \emph{code collapse}.
Existing methods typically address these desiderata separately.
By contrast, the present study introduces a simple and intuitive regularization that promotes both simultaneously by minimizing the distance between each simplex vertex and its $K$-nearest smoothed code-selectors.
Representative benchmarks
on image encoding
demonstrate that the proposed method achieves more effective codebook utilization and improves performance over prior approaches.
\end{abstract}

\section{Introduction}
\label{sec:intro}

Vector quantization is a method for discretizing a continuous vector space \citep{Gray84,vandenOord+17_VQVAE}.
It maps each vector in the continuous space to the nearest element of a finite set of representative vectors (a.k.a. a \emph{codebook}).
The resulting discrete representations are easier to manipulate and interpret than the original continuous forms,
and have proved effective across diverse applications, including image generation \citep{Esser+21,Ramesh+21,Rombach+22,Yu+22_Parti}, speech recognition \citep{Baevski+20,Baevski+20_wav2vec2.0}, and speech/music generation \citep{HadjeresCrestel20,Dhariwal+20_Jukebox,Zhang+23,Chen+25}.

When integrated into deep neural networks, however, vector quantization introduces a fundamental challenge:
quantization is a non-differentiable operation that blocks gradient backpropagation \citep{vandenOord+17_VQVAE}.
Accordingly, some approximation is required to enable learning through quantization.

One effective workaround is to \emph{smooth} vector quantization \citep{Jang+17}.
The selection of a codebook vector can be expressed as multiplying the codebook matrix by its corresponding onehot vector $\mathbf{e}_m:=(0,\dots,1,\dots,0)^{\mathsf{T}}$, whose nonzero entry at $m$ indexes the chosen vector.
\emph{Smoothed} vector quantization relaxes this onehot vector to lie within the simplex $\Delta^{M-1} := \{ (p_1,\dots,p_M) \in \mathbb{R}^M \mid \sum_{m=1}^{M} p_m = 1, p_m \geq 0 (\forall m)\}$, where $M$ denotes the number of codebook vectors.
Consequently, differentiable mappings (e.g., softmax) become available and can be incorporated into neural networks.

\begin{figure}
	\centering
	\begin{subcaptiongroup}
		\phantomcaption\label{fig:simplex}
		\phantomcaption\label{fig:entropy-of-mean}
		\phantomcaption\label{fig:knnl2}
	\end{subcaptiongroup}
	\includegraphics[width=\columnwidth]{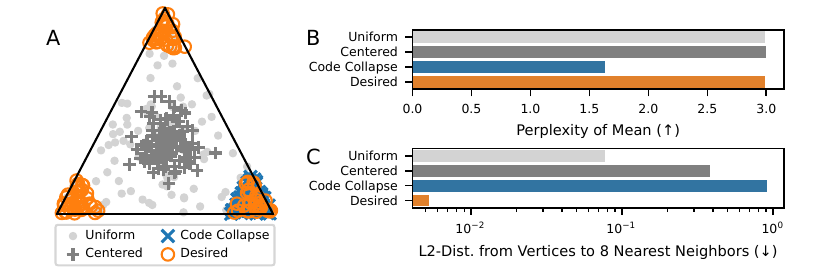}
	\caption{(\subref{fig:simplex}) Four different distributions on the simplex $\Delta^{3-1}$.
	For effective smoothed vector quantization, samples should be concentrated near the vertices of the simplex (i.e., onehot-like vectors; orange), rather than centered (dark gray) or uniformly spread across the simplex (light gray).
	At the same time, each vertex must have some samples in its neighborhood to avoid code collapse (blue).
	(\subref{fig:entropy-of-mean}) Maximizing the perplexity of the sample mean \citep{Baevski+20_wav2vec2.0} penalizes code collapse but cannot discriminate among the other three distributions.
	(\subref{fig:knnl2}) The proposed $K$-nearest neighbor (KNN) distance minimization ($K=8$) favors the desired vertex-concentrated distribution while also preventing code collapse.
	}
\end{figure}

To successfully approximate onehot code-selection vectors, smoothed vectors must be distributed around the vertices of the simplex (orange ``\textsf{o}'' in Figure~\ref{fig:simplex}).
At the same time, they should not concentrate around a few vertices, leaving other codebook entries unused (blue ``\textsf{x}'').
This latter issue---called \emph{code collapse}---has been identified as a major challenge in vector quantization \citep{Dieleman+18,Baevski+20_wav2vec2.0,Dhariwal+20_Jukebox,Fifty+25}.
A widely adopted workaround for code collapse (in smoothed quantization) is to introduce an auxiliary learning objective that maximizes the entropy or perplexity of the \emph{mean} of the smoothed code-selection vectors \citep[see \textsection\ref{sec:regularization} for the formal definition]{Dieleman+18,Baevski+20_wav2vec2.0}.
However, maximizing the entropy/perplexity of the mean can be achieved by various distributions, not only the desired vertex-neighboring ones.
For example, both uniform and center-concentrated distributions have their mean at the simplex center (Figure~\ref{fig:simplex}), which also maximizes this objective (Figure~\ref{fig:entropy-of-mean}).
Accordingly, an additional mechanism is needed to tighten the smoothing (e.g., by adjusting the temperature parameter of the (Gumbel-)softmax; see \textsection\ref{sec:gumbel}).

Beyond this standard approach, however, there exists a simple and unified strategy for simultaneously tightening smoothed quantization and maximizing codebook usage:
\emph{Why not directly encourage clustering around all simplex vertices?}
Specifically, minimizing the distance between each simplex vertex and its $K$-nearest neighbors (KNNs) satisfies both desiderata at once (Figure~\ref{fig:knnl2}).
The present study investigates this intuitive yet underexplored approach, comparing it against existing alternatives on representative benchmarks.
The results indicate that the proposed method enables the exhaustive usage of the entire codebook, even when other approaches suffer from code collapse.

The remainder of this paper is organized as follows.
\textsection\ref{sec:review} reviews related studies on vector quantization.
\textsection\ref{sec:methods} introduces the proposed method,
which is then evaluated on representative benchmarks in \textsection\ref{sec:experiments}.
Finally, \textsection\ref{sec:discussions} discusses the results and the limitations of the proposed method.

\section{Related Studies}
\label{sec:review}

The central challenge of vector quantization lies in its non-differentiability, which disrupts the backpropagation of gradients in neural networks.
To address this, either the gradient computation (backward pass) or the quantization itself (forward pass) must be approximated.

\subsection{Approximation in the Backward Pass (Gradient Estimation)}
\label{sec:vq}

One line of work retains the original non-differentiable quantization in the forward pass but replaces the gradient computation in the backward pass.
Let $\mathbf{z} \in \mathbf{R}^D$ denote a pre-quantized feature vector, and $\{\mathbf{q}_1,\dots, \mathbf{q}_M\} \subset \mathbf{R}^D$ the set of quantized vectors.
Quantization maps $\mathbf{z}$ to its ``closest'' codebook entry according to a distance metric $\mathcal{D}$:
i.e., $\mathbf{z} \mapsto \mathbf{q}_{\iota(\mathbf{z})}$ where $\iota(\mathbf{z}) := \operatorname{argmin}_{m} \mathcal{D}(\mathbf{z}, \mathbf{q}_m)$.
Consequently, the partial derivatives $\frac{\partial q_{\iota(\cdot),i}}{\partial z_j}$ are ill-defined and must be approximated.

A canonical approximation---known as the \emph{straight-through estimation} (STE)---replaces the ill-defined Jacobian with the identity matrix \citep{vandenOord+17_VQVAE}:
\begin{align}
	\frac{\partial q_{\iota(\cdot),i}}{\partial z_j}
		\approx
			\begin{cases}
				1&
					i=j\\
				0&
					\text{otherwise}
			\end{cases}
\end{align}
Algorithmically, the STE is implemented using the $\mathtt{detach}$ operation, which excludes its argument from gradient computation:
\begin{align}
	\operatorname{STE}(\mathbf{q}_{\iota(\mathbf{z})}, \mathbf{z})
		&:=
			\mathbf{q}_{\iota(\mathbf{z})} + \mathbf{z} - \mathtt{detach}(\mathbf{z})
			\label{eq:vq_STE}
\end{align}
Here, $z - \mathtt{detach}(z)$ evaluates to zero, while its gradient with respect to $z$ is the identity.

More recently, \citet{Fifty+25} proposed an alternative gradient approximation and demonstrated its empirical superiority over the STE:
\begin{align}
	\textsc{RE}(\mathbf{q}_{\iota(\mathbf{z})}, \mathbf{z})
		&:=
			\mathtt{detach}\left(
				\frac{\| \mathbf{q}_{\iota(\mathbf{z})} \|}{\| \mathbf{z} \|}
				R
			\right)
			\mathbf{z}
			\label{eq:vq_rot}
\end{align}
where $R$ is the rotation matrix aligning $\mathbf{z}$ to $\mathbf{q}_{\iota(\mathbf{z})}$,%
\footnote{
	The rotation matrix is given by $R = I - 2\hat{\mathbf{r}}\hat{\mathbf{r}}^{\mathsf{T}} + 2\hat{\mathbf{q}}_{\iota(\mathbf{z})}\hat{\mathbf{z}}^{\mathsf{T}}$, where $\hat{\mathbf{v}} := \mathbf{v} / \| \mathbf{v} \|$ is the L2-normalization of vector $\mathbf{v}$, and $\mathbf{r} := \hat{\mathbf{q}}_{\iota(\mathbf{z})} + \hat{\mathbf{z}}$.
}
and $\frac{\| \mathbf{q}_{\iota(\mathbf{z})} \|}{\| \mathbf{z} \|}$ rescales the rotated vector to match the amplitude of $\mathbf{q}_{\iota(\mathbf{z})}$.
In this formulation, the Jacobian of the quantization is approximated by the scaled rotation matrix:
\begin{align}
	\frac{\partial q_{\iota(\cdot),i}}{\partial z_j}
		\approx
			\frac{\| \mathbf{q}_{\iota(\mathbf{z})} \|}{\| \mathbf{z} \|}
				R
\end{align}

\subsection{Approximation in the Forward Pass (Smoothing)}
\label{sec:gumbel}

An alternative approach approximates the forward quantization itself.
Using the onehot representation $\mathbf{e}_{m}$ of the code index $m$, quantization can be expressed as:
\begin{align}
	\mathbf{q}_{\iota(\mathbf{z})}
		&=
			Q\mathbf{e}_{\iota(\mathbf{z})}
\end{align}
where $Q := (\mathbf{q}_1, \dots, \mathbf{q}_M)$.
\emph{Smoothed} quantization extends the possible range of $\mathbf{e}_{m}$ to the simplex $\Delta^{M-1}$.
As noted in \textsection\ref{sec:intro}, effective learning requires smoothed code-selectors $\mathbf{p} \in \Delta^{M-1}$ to concentrate near the vertices of the simplex (i.e., $\mathbf{p} \approx \mathbf{e}_{m}$ for some $m$).

A widely studied instance of smoothed quantization is Gumbel-softmax sampling \citep{Jang+17}.
Given assignment probabilities $\pi_m$ of $\mathbf{z}$ to the $m$-th code---typically log-proportional to their dot-product $\mathbf{q}_m^{\mathsf{T}} \mathbf{z}$---categorical sampling can be implemented using Gumbel samples $g_m = -\log(-\log u_m)$ with $u_m \sim \operatorname{Uniform}(0,1)$:
\begin{align}
	&
		\iota(\mathbf{z}) \sim \operatorname{Categorical}(\pi_1,\dots,\pi_M)
		\nonumber
		\\
	\Leftrightarrow
	&
		\iota(\mathbf{z}) = \underset{m}{\operatorname{argmax}} (g_m + \log \pi_m)
		\label{eq:gumbel-max}
\end{align}
Replacing $\operatorname{argmax}$ above with $\operatorname{softmax}$ yields a smoothed quantization:
\begin{align}
	p_{m}
		=&
			\frac{\exp\left( \left( g_m + \log \pi_m \right) / \tau \right)
			}{\sum_{m'} \exp\left( \left( g_{m'} + \log \pi_{m'} \right) / \tau \right)}
			\label{eq:gumbel-softmax}
\end{align}
where lowering the temperature parameter $\tau$ produces a tighter approximation of categorical sampling.

The Gumbel-softmax sampling can also be combined with hard quantization using the STE:
\begin{align}
	\operatorname{STE}(\mathbf{e}_{\iota(\mathbf{z})}, \mathbf{p})
		=
			\mathbf{e}_{\iota(\mathbf{z})}
			+ \mathbf{p}
			- \mathtt{detach}(\mathbf{p})
\end{align}

\subsection{Regularization}
\label{sec:regularization}

Beyond approximation strategies, vector quantization also requires auxiliary regularization losses to ensure effective training.
For example, the STE alone does not guarantee alignment between pre-quantized features and codebook entries.
This alignment is instead fostered by the following regularization loss, $\mathcal{L}_{\text{reg}}$ \citep{vandenOord+17_VQVAE,Fifty+25}:
\begin{align}
	\mathcal{L}_{\text{total}}
		&=
			\mathcal{L}_{\text{main}}
			+ \mathcal{L}_{\text{reg}}
			\label{eq:total_loss}
			\\
	\mathcal{L}_{\text{reg}}
		&=
	\mathcal{L}_{\text{hard}}
		:=
			N^{-1} \sum_{i=1}^N
			\left(
				\beta \underbracket{\| \mathbf{z}_i - \mathtt{detach}(\mathbf{q}_{
					m_i
					}) \|_2^2}_{\text{Commitment Loss}}
			\right.
			\left.
				+
				\underbracket{\| \mathtt{detach}(\mathbf{z}_i) - \mathbf{q}_{
					m_i
					} \|_2^2}_{\text{Codebook Loss}}
			\right)
			\label{eq:vq-loss}
\end{align}
where $\mathcal{L}_{\text{main}}$ is the primary task loss (e.g., L2 regression in autoencoding), and $\beta>0$ is a weighting hyperparameter.
As defined in Equation~\ref{eq:vq-loss}, $\mathcal{L}_{\text{hard}}$ consists of two components.
The first term (known as the \emph{commitment loss}) aligns each pre-quantized feature $\mathbf{z}_i$ in a batch ($i=1,\dots,N$) with its nearest codebook entry $\mathbf{q}_{
	m_i
	}$.
The second term (called the \emph{codebook loss}) moves each codebook vector $\mathbf{q}_m$ toward the centroid of its assigned features whose nearest neighbor is $\mathbf{q}_m$ (i.e., $\{\mathbf{z}_i: \operatorname{argmin}_{m'} \| \mathbf{z}_i - \mathbf{q}_{m'} \| = m \}$).

It should be noted, however, that the codebook loss in Equation~\ref{eq:vq-loss} does not inherently prevent code collapse.
Some codebook vectors may never serve as the nearest neighbor of any pre-quantized feature and therefore receive no updates \citep{Zhu+25}.
Accordingly, previous studies have resorted to additional workarounds; for example, unused codebook vectors may be reset to the positions of pre-quantized features \citep{Dhariwal+20_Jukebox}.
More recently, \citet{Zhu+25} proposed another remedy called \emph{SimVQ}, which reparameterizes the codebook $Q$ as the product of a randomly frozen matrix $Q' \in \mathbb{R}^{M \times D}$ and a learnable matrix $W \in \mathbb{R}^{D \times D}$ (i.e., $Q = Q'W$).
In this formulation, all codebook vectors share learnable parameters with one another ($\mathbf{q}_m = q'_{m,1} \mathbf{w}_1 + \cdots + q'_{m,D} \mathbf{w}_D$), so updates to one vector propagate to the others.


Similarly, smoothed quantization also requires auxiliary regularization to avoid code collapse.
A widely adopted option is to maximize the normalized perplexity of the mean assignment probability
\citep{Dieleman+18,Baevski+20_wav2vec2.0}:
\begin{align}
	\mathcal{L}_{\text{reg}}
		&=
			\mathcal{L}_{\text{ppl}}
			:=
			-\frac{\exp\left(
				-\sum_{m} \bar{\pi}_m \log \bar{\pi}_m
			\right)}{M}
			\label{eq:ppl}
			\\
	\bar{\pi}_m
		&:=
			N^{-1} \sum_{i=1}^N \pi_{i,m}
\end{align}
As noted in \textsection\ref{sec:intro}, this perplexity-based regularization does not promote the onehotness of $\boldsymbol{\pi}_i$, failing to distinguish onehot-like samples from uniform or centered ones (Figure~\ref{fig:entropy-of-mean}).
Onehotness can instead be induced by annealing the temperature parameter of the Gumbel-softmax sampling ($\tau \to 0$).
However, manually scheduling this annealing is empirically challenging.
The next section therefore proposes an alternative regularization loss that automatically encourages onehotness within the framework of gradient-based learning, while simultaneously preventing code collapse.

\section{Methods}
\label{sec:methods}

As noted in previous sections, ideal smoothed code-selection vectors $\mathbf{p} \in \Delta^{M-1}$ are distributed near the vertices of the simplex (Figure~\ref{fig:simplex}).
Moreover, each vertex should have at least some smoothed code-selectors in its neighborhood; otherwise, the quantization suffers from code collapse.
A simple way to achieve these objectives simultaneously is to impose a loss penalizing the deviation of the KNNs from each vertex (Figure~\ref{fig:knnl2}):%
\footnote{
	A possible alternative formalization and its limitations are discussed in Appendix~\ref{sec:dirichlet}.
}
\begin{align}
	\mathcal{L}_{\text{reg}}
		&=
			(MK)^{-1}
			\sum_{m=1}^{M}
			\sum_{k=1}^{K}
			\mathcal{D}(\mathbf{e}_{m}, \mathbf{p}^{(m,k)})
			\label{eq:knn-reg}
\end{align}
where $\mathbf{p}^{(m,k)}$ 
denotes the $k$-th nearest neighbor of the simplex vertex (onehot vector) $\mathbf{e}_m$ according to a distance/deviation metric $\mathcal{D}$.
Two options for $\mathcal{D}$ are considered in this study: the squared L2 distance, $\| \mathbf{e}_{m} - \mathbf{p}^{(m,k)} \|_2^2$, and cross-entropy, $- \log p^{(m,k)}_m$.

The proposed regularization no longer requires Gumbel-softmax sampling;
smoothed code-selectors can be obtained directly as $\mathbf{p} = \boldsymbol{\pi} = \operatorname{softmax}(Q^{\mathsf{T}} \mathbf{z})$.%
\footnote{
	The exact implementation of $\mathbf{p}$ and $\boldsymbol{\pi}$ involves normalization and rescaling; see Appendix~\ref{sec:smooth_detail} for details. 
}
At the same time, the proposed regularization is fully compatible with Gumbel-softmax sampling; one can simply replace $\mathbf{p}$ (Gumbel-softmax samples) in Equation~\ref{eq:knn-reg} with $\boldsymbol{\pi}$ (assignment probabilities).

During inference, hard quantization is applied by taking $\operatorname{argmax}_m p_m$ in the onehot representation.
In the following section, both the deterministic and stochastic approaches are evaluated on representative benchmarks.

\section{Experiments}
\label{sec:experiments}

The proposed regularization for smoothed vector quantization was benchmarked on representative tasks of discrete autoencoding (\textsection\ref{sec:ae}) and more sophisticated adversarial training (VQGAN; \textsection\ref{sec:vqgan_image}).

\subsection{Discrete Autoencoding}
\label{sec:ae}

The first benchmark assessed the proposed regularization in the context of discrete autoencoding on the ImageNet dataset \citep{Deng+09}.
Input images were convolutionally encoded into latent feature maps, whose pixels were subsequently quantized \citep{Esser+21,Fifty+25}.
A decoder convolutional network reconstructed the input images from these quantized feature maps, and the entire model was trained to minimize the L2 reconstruction loss ($L_{\text{main}}$ in Equation~\ref{eq:total_loss}).
Further details of the network architecture and training setup are provided in Appendix~\ref{sec:ae_detail}.

\begin{table*}
	\centering
	\caption{Performance of discrete autoencoding on the ImageNet validation set. Reported metrics are codebook usage and reconstruction quality scores: root mean squared error (rMSE), Fr\'{e}chet Inception Distance (FID), Inception Score (IS), and Structural Similarity Index Measure (SSIM). The proposed method is denoted as ``KNN-L2/CE''.
	The baseline methods are abbreviated as follows:
	``Cos'' denotes cosine-distance quantization (contrasted with Euclidean distance, 'Euclid'). ``HG'' and ``SG'' denote Hard-Gumbel and Soft-Gumbel, i.e., Gumbel-softmax sampling with and without straight-through hard quantization in the forward pass, respectively.
	Best scores across all methods are highlighted in boldface, while underlined values indicate the best-performing value of $K/4$.
	}
	\label{tab:score_ae}
	{\fontsize{6.5}{7.5}\selectfont
	\setlength{\tabcolsep}{0.45em}
	\begin{tabular}{llcrrrrrrrrrrrr}
	\toprule
		&&&
			\multicolumn{12}{c}{Feature\,Map\,Size; Codebook\,Size}
			\\
		&&&
			\multicolumn{4}{c}{16$\times$16$\times$32; 1024}
			&
			\multicolumn{4}{c}{64$\times$64$\times$3; 8196}
			&
			\multicolumn{4}{c}{64$\times$64$\times$64; 8196}
			\\
		\cmidrule(l{0.4em}r{0.4em}){4-7}
		\cmidrule(l{0.4em}r{0.4em}){8-11}
		\cmidrule(l{0.4em}r{0.4em}){12-15}
		\multicolumn{2}{l}{Method}&
		$K/4$&
			Usage\,($\uparrow$)&
				rMSE\,($\downarrow$)&
					FID\,($\downarrow$)&
						IS\,($\uparrow$)&
			Usage\,($\uparrow$)&
				rMSE\,($\downarrow$)&
					FID\,($\downarrow$)&
						IS\,($\uparrow$)&
			Usage\,($\uparrow$)&
				rMSE\,($\downarrow$)&
					FID\,($\downarrow$)&
						IS\,($\uparrow$)
							\\
\midrule
\multirow[c]{3}{*}[-.75em]{\rotatebox{90}{STE}} & Euclid & --- & 4.5\% & 0.108 & 101.84 & 10.67 & \bftab 100.0\% & \bftab 0.045 & \bftab 10.41 & \bftab 142.12 & 1.7\% & 0.067 & 25.65 & 82.61 \\
\cmidrule{2-15}
 & Cos & --- & 3.0\% & 0.102 & 108.09 & 10.03 & 70.9\% & 0.053 & 18.13 & 107.38 & 1.7\% & 0.054 & 20.08 & 101.74 \\
\cmidrule{2-15}
 & SimVQ & --- & \bftab 100.0\% & \bftab 0.091 & 86.94 & 15.93 & \bftab 100.0\% & 0.046 & 10.82 & 141.17 & \bftab 100.0\% & 0.045 & 12.54 & 134.84 \\
\midrule
\multirow[c]{2}{*}[-.5em]{\rotatebox{90}{RE}} & Euclid & --- & 3.1\% & 0.123 & 127.40 & 7.40 & 78.8\% & 0.046 & 14.04 & 123.30 & 0.2\% & 0.077 & 42.12 & 51.86 \\
\cmidrule{2-15}
 & Cos & --- & 2.8\% & 0.113 & 130.47 & 7.69 & 99.5\% & 0.052 & 19.49 & 102.55 & 2.9\% & 0.050 & 20.11 & 101.75 \\
\midrule
\multirow[c]{2}{*}[-.5em]{\rotatebox{90}{HG}} & PPL & --- & \bftab 100.0\% & 0.094 & 94.81 & 12.26 & \bftab 100.0\% & 0.051 & 24.41 & 87.22 & \bftab 100.0\% & \bftab 0.044 & 14.77 & 124.10 \\
\cmidrule{2-15}
 & \bftab KNN-CE & 1 & \bftab 100.0\% & 0.099 & 77.21 & 18.41 & \bftab 100.0\% & 0.061 & 18.26 & 111.14 & \bftab 100.0\% & 0.045 & 4.56 & 179.47 \\
\midrule
\multirow[c]{2}{*}[-.5em]{\rotatebox{90}{SG}} & PPL & --- & 55.0\% & 0.224 & 161.12 & 3.76 & \bftab 100.0\% & 0.049 & 14.34 & 121.54 & 49.8\% & 0.320 & 292.60 & 1.85 \\
\cmidrule{2-15}
 & \bftab KNN-CE & 1 & \bftab 100.0\% & 0.153 & 161.97 & 6.16 & \bftab 100.0\% & 0.063 & 19.24 & 106.64 & \bftab 100.0\% & 0.048 & 4.62 & 180.80 \\
\midrule
\multirow[c]{9}{*}{\rotatebox{90}{Softmax}} & PPL & --- & \bftab 100.0\% & 0.250 & 260.30 & 2.75 & 81.6\% & 0.164 & 57.66 & 49.21 & 98.4\% & 0.149 & 149.05 & 13.71 \\
\cmidrule{2-15}
 & \multirow[c]{4}{*}{\bftab KNN-L2} & 1 & \bftab 100.0\% & 0.096 & 74.53 & 22.33 & \bftab 100.0\% & 0.060 & 16.16 & 119.58 & \bftab 100.0\% & 0.092 & 14.38 & 117.34 \\
 &  & 2 & \bftab 100.0\% & 0.098 & 75.07 & 20.38 & \bftab 100.0\% & 0.061 & 15.78 & 116.06 & \bftab 100.0\% & 0.070 & 7.91 & 151.99 \\
 &  & 4 & \bftab 100.0\% & 0.102 & 78.16 & 18.28 & \bftab 100.0\% & 0.055 & \underline{13.15} & \underline{132.91} & \bftab 100.0\% & \underline{0.051} & \underline{5.96} & \underline{168.61} \\
 &  & 8 & 99.9\% & \underline{0.092} & \bftab 71.81 & \bftab 22.71 & \bftab 100.0\% & \underline{0.054} & 16.88 & 114.00 & \bftab 100.0\% & 0.056 & 8.71 & 152.68 \\
\cmidrule{2-15}
 & \multirow[c]{4}{*}{\bftab KNN-CE} & 1 & \bftab 100.0\% & \underline{0.098} & \underline{75.16} & \underline{21.10} & \bftab 100.0\% & 0.060 & \underline{18.59} & \underline{110.88} & \bftab 100.0\% & \bftab 0.044 & \bftab 2.86 & \bftab 194.38 \\
 &  & 2 & \bftab 100.0\% & 0.102 & 95.95 & 13.69 & \bftab 100.0\% & 0.061 & 19.71 & 107.15 & \bftab 100.0\% & 0.046 & 3.55 & 187.71 \\
 &  & 4 & \bftab 100.0\% & 0.200 & 227.82 & 3.28 & \bftab 100.0\% & 0.061 & 21.58 & 99.60 & \bftab 100.0\% & 0.050 & 6.12 & 166.37 \\
 &  & 8 & \bftab 100.0\% & 0.327 & 260.67 & 1.96 & \bftab 100.0\% & \underline{0.059} & 18.92 & 107.23 & \bftab 100.0\% & 0.053 & 6.37 & 162.22 \\
\bottomrule
\end{tabular}
	}
\end{table*}

Three feature-map configurations were considered.
Two followed the configurations used in previous studies \citep{Esser+21,Fifty+25}, with $H \times W \times C = 16 \times 16 \times 32$ ($M = 1024$) and
$64 \times 64 \times 3$ ($M=8196$).
For the latter configuration, the channel dimensionality was further increased to $C = 64$, resulting in feature maps of size $64 \times 64 \times 64$ while retaining $M = 8196$.

Each model was trained using four GPUs in parallel. Accordingly, the $K/4$-nearest neighbors of each simplex vertex were computed independently on each GPU, where $K/4 \in \{1, 2, 4, 8\}$.%
\footnote{
	The value of $K$ was upper-bounded at $8 \times 4$ by available computational resources;
	with a maximum batch size of $64$, the total number of latent pixels was $64 \times 64 \times 64 = 8 \times 4 \times 8196$, allowing only $8 \times 4$ neighbors per vertex of $\Delta^{8196-1}$.
	This implementation constraint is further discussed in \textsection\ref{sec:vqgan_image}.
}
Crucially, performing the $K$-nearest neighbor search locally on each device avoids expensive multi-GPU communication (such as \texttt{all\_gather} operations in PyTorch), which would otherwise be required to share assignment probabilities across GPUs. This design makes the proposed regularization highly parallelizable and computationally efficient (see Appendix~\ref{sec:compute-load} for a detailed analysis of wall-clock time).

The weight $\beta$ on the commitment loss (Equation~\ref{eq:vq-loss}) for STE-based methods (including SimVQ) and rotational gradient estimation was set to 1.0, following \citet{Fifty+25} and \citet{Zhu+25}.

\begin{table*}
	\centering
	\caption{Tightness of softmax-based smoothing (without Gumbel sampling), measured by the individual perplexity of smoothed code-selection vectors, $\exp(-\sum_{m=1}^{M} p_m \log p_m)$. Reported values are the 75th, 90th, and 99th percentiles, as well as the maximum, computed across all feature-map pixels in the ImageNet validation set.}
	\label{tab:tightness}
	{\fontsize{7}{8}\selectfont
	\setlength{\tabcolsep}{0.45em}
\begin{tabular}{lcrrrrrrrrrrrr}
\toprule
 &  & \multicolumn{12}{c}{Feature\,Map\,Size; Codebook\,Size} \\
 & & \multicolumn{4}{c}{16$\times$16$\times$32; 1024} & \multicolumn{4}{c}{64$\times$64$\times$3; 8192} & \multicolumn{4}{c}{64$\times$64$\times$64; 8192} \\
\cmidrule(l{0.4em}r{0.4em}){3-6}
\cmidrule(l{0.4em}r{0.4em}){7-10}
\cmidrule(l{0.4em}r{0.4em}){11-14}
Method& $K/4$& 75\% & 90\% & 99\% & Max & 75\% & 90\% & 99\% & Max & 75\% & 90\% & 99\% & Max\\
\midrule
PPL & --- & 910.40 & 933.12 & 963.29 & 995.91 & 7559.89 & 7561.99 & 7563.10 & 7565.02 & 8186.32 & 8186.40 & 8186.55 & 8186.76 \\
\midrule
\multirow{4}{*}{KNN-L2} & 1 & 1.17 & 1.59 & 2.36 & 6.26 & 1.00 & 1.00 & 1.00 & 4.00 & 9.63 & 18.10 & 51.04 & 123.62 \\
 & 2 & 1.00 & 1.08 & 1.86 & 27.37 & 1.00 & 1.00 & 1.06 & 4.65 & 3.85 & 6.36 & 16.53 & 43.40 \\
 & 4 & 1.00 & 1.06 & 1.86 & 62.99 & 1.00 & 1.00 & 1.00 & 4.61 & 1.06 & 1.06 & 1.06 & 1.07 \\
 & 8 & 1.00 & 1.10 & 1.90 & 120.68 & 3.33 & 3.87 & 5.22 & 12.03 & 1.06 & 1.06 & 1.06 & 1.06 \\
\midrule
\multirow{4}{*}{KNN-CE} & 1 & 1.18 & 1.60 & 2.38 & 6.28 & 3.96 & 4.51 & 5.81 & 11.96 & 1.21 & 1.21 & 1.21 & 1.22 \\
 & 2 & 1.23 & 1.72 & 2.73 & 8.07 & 4.04 & 4.66 & 6.16 & 11.77 & 1.18 & 1.18 & 1.18 & 1.19 \\
 & 4 & 1.42 & 1.78 & 2.38 & 5.45 & 4.16 & 4.86 & 6.57 & 13.04 & 1.21 & 1.22 & 1.22 & 1.23 \\
 & 8 & 2.38 & 2.80 & 3.71 & 7.78 & 4.46 & 5.31 & 7.39 & 13.81 & 1.43 & 1.43 & 1.44 & 1.46 \\
\bottomrule
\end{tabular}
	}
\end{table*}

Table~\ref{tab:score_ae} reports the codebook utilization and reconstruction performance---including root mean squared error (rMSE), Inception Score \citep[IS;][]{Salimans+16}, and Fr\'{e}chet Inception Distance \citep[FID;][]{Heusel+17}---for the proposed method (``KNN-L2/CE'') and the baselines.
Reconstruction quality deteriorated substantially when softmax-smoothed quantization (i.e., without Gumbel randomness) was combined with the perplexity-based regularization (``PPL''; Equation~\ref{eq:ppl}).
This degradation arises from a mismatch between soft quantization during training and hard quantization at inference; the smoothed code-selectors deviated from the simplex vertices, as reflected by their high individual perplexity, $\exp(-\sum_{m=1}^{M} p_m \log p_m)$ (Table~\ref{tab:tightness}).
These results support the argument made in the Introduction that perplexity-based regularization alone does not promote tight smoothing.

Perplexity-based regularization was effective only when combined with the Gumbel-softmax sampling,
and onehot quantization in the forward pass via STE was necessary to ensure full codebook utilization (``\underline{H}ard-\underline{G}umbel'').
Without hard quantization, code collapse was not prevented at higher channel dimensionalities ($C \geq 32$; ``\underline{S}oft-\underline{G}umbel'').

By contrast, the proposed KNN-based regularization consistently prevented code collapse and achieved near-complete codebook utilization without resorting to Gumbel-softmax sampling.
Reconstruction performance was also superior to, or competitive with, the Gumbel-softmax plus perplexity baseline across all feature map sizes.
Moreover, eliminating the Gumbel-softmax sampling reduces the runtime as well
(see Appendix~\ref{sec:compute-load}).

The choice of deviation metric (L2 distance vs. cross-entropy, CE) did not exhibit a consistent global advantage in reconstruction quality.
%
%
%
However, cross-entropy was additionally favorable in terms of computational efficiency.
Its performance was highest when using the minimum number of neighbors ($K/4=1$) for all the evaluation metrics (underlined entries in Table~\ref{tab:score_ae}) except rMSE under the $64 \times 64 \times 3$ setting; increasing $K$ yielded no further gains.
Although most smoothed code-selectors were formally left unregularized under this setting, they nonetheless achieved low individual perplexity after training (Table~\ref{tab:tightness}), indicating a tight approximation to hard quantization.
While the exact mathematical mechanism remains an open question, these consistent empirical findings suggest that anchoring only the single closest sample is sufficient to drive the entire distribution toward the simplex vertices.
By contrast, minimizing the L2 distance to only a single nearest neighbor per GPU led to performance degradation for feature maps of size $64 \times 64 \times 64$.
Although reconstruction performance alone does not favor either metric consistently, this asymmetry in the reliability of small-$K$ behavior matters for scalability,
as increasing the number of neighbors requires a larger batch size (see \textsection\ref{sec:vqgan_image} for details).

\begin{figure*}
	\centering
	\includegraphics[width=\textwidth]{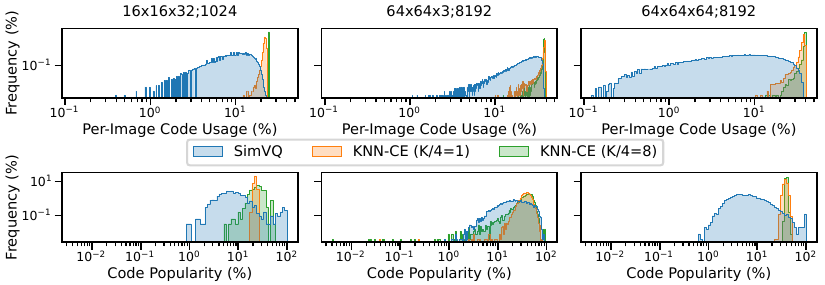}
	\caption{(Top) Histogram of per-image codebook utilization on the validation dataset, defined as the percentage of codes used at least once within each image. (Bottom) Histogram of code popularity, defined as the proportion of validation images in which each code occurs.}
	\label{fig:code-use-hist}
\end{figure*}

Finally, both STE and rotational gradient estimation (RE) exhibited severe code collapse at higher channel dimensionalities ($C \geq 32$), reaffirming prior findings that reducing channel dimensionality is critical for stable training \citep{Yu+22,Yu+24}.
SimVQ was more robust, maintaining full codebook utilization under all settings.
Looking at individual images, however, more than 5\% of them were encoded by using only 1\% or a smaller subset of the codes under the $64 \times 64 \times 64$ condition (Figure~\ref{fig:code-use-hist}).
This lack of code diversity per image resulted in performance degradation compared to the low-dimensional setting ($64 \times 64 \times 3$), 
even though the best scores (achieved by the KNN-CE with $K/4=1$) improved by increasing the channel dimensionality.




\subsection{VQGAN and Product Quantization}
\label{sec:vqgan_image}

A primary limitation of the proposed method is the empirical upper bound on the codebook size $M$.
The KNN-based regularization identifies $K$ nearest smoothed code-selectors ($\mathbf{p}$ or $\boldsymbol{\pi}$) per simplex vertex (on each GPU).
Consequently, each batch must contain at least $KM$ latent pixels/frames.
This requirement limits the scalability of the codebook size $M$ and, consequently, the expressiveness of the quantized representations under standard VRAM constraints.

The present section introduces a possible workaround for this problem and demonstrate its effectiveness based on an advanced quantization task requiring a larger codebook size than naive autoencoding.
Specifically, the proposed method is combined with \emph{product quantization}, which achieves diverse discrete representations by combining multiple codebooks \citep{Jegou+11,Baevski+20_wav2vec2.0}.
By splitting a codebook matrix of size $M \times D$ into $G$ smaller codebooks of size $M \times D/G$, product quantization can represent $M^G$ distinct quantized vectors.
Thus, a huge effective vocabulary (of size $M^G$) is achievable without increasing the size $M$ of individual codebooks, thereby circumventing the scalability constraint of the proposed method.

To demonstrate the effectiveness of product quantization, the proposed method was evaluated 
on the VQGAN task \citep{Esser+21}.
While the network architecture remains identical to the discrete autoencoder (\textsection\ref{sec:ae}), VQGAN uses additional learning objectives besides the pixel-wise error minimization in order to improve the quality of reconstructed images (see Appendix~\ref{sec:vqgan_detail} for details).
To fully exploit these advanced objectives, however, increasing the effective codebook size and thereby the granularity of discrete latent representations is crucial.
Thus, VQGAN is a representative benchmark for evaluating the combination of product quantization and KNN-based regularization.

Following the original study \citep{Esser+21},
quantization was performed on
the latent feature map of size $16 \times 16 \times 256$,
derived from the original input image of size $256 \times 256 \times 3$.
Product quantization was implemented with $G=8$ groups and a codebook size $M = 1024$, thereby splitting the latent channels into $D/G = 256/8=32$ while increasing the effective vocabulary to $M^G = 1024^8$.

\begin{table*}
	\centering
	\caption{Performance of VQGAN on the ImageNet validation set. Reported metrics are codebook usage and reconstruction quality scores.
	Best scores across all methods are highlighted in boldface.
	For the single-codebook STE/RE Euclid baselines ($G=1$),
	scores from the original publications are reported,
	since attempts to reproduce these results were unsuccessful.\protect\footnotemark\ 
	All product-quantized configurations ($G=8$), including these baselines, were newly implemented and evaluated.
	}
	\label{tab:score_vqgan}
	{\fontsize{7.5}{8.5}\selectfont
	\setlength{\tabcolsep}{0.5em}
	\begin{tabular}{llrrcrr}
	\toprule
		Method&
		Evaluated in&
		$D/G$&
		$M^G$&
			Usage\,\%\,($\uparrow$)&
					FID\,($\downarrow$)&
						IS\,($\uparrow$)
							\\
	\midrule
		\multirow{3}{*}[-0.5em]{STE Euclid}
		&
		\multirow{2}{*}{\citet{Esser+21}}
		&
		$256/1$&
		$1024^1$&
			---&
					7.94&
						---
							\\
		&
		&
		$256/1$&
		$16384^1$&
			---&
					4.98&
						---
							\\
		\cmidrule{2-7}
		&
		This\,Work
		&
		$256/8$&
		$1024^8$&
			0.1--0.2&
					398.52&
						1.00
							\\
		\midrule
		\multirow{2}{*}[-0.4em]{RE Euclid}
		&
		\citet{Fifty+25}
		&
		$256/1$&
		$1024^1$&
			98&
					4.6&
						146.5
							\\
		\cmidrule{2-7}
		&
		This\,Work
		&
		$256/8$&
		$1024^8$&
			0.2--0.8&
					506.60&
							1.00\\
		\midrule
		\multirow{3}{*}{SimVQ}
		&
		\multirow{3}{*}{This\,Work\footnotemark}
		&
		$32/1$&
		$1024^1$&
			1.0&
					436.34&
							1.00\\
		&
		&
		$128/1$&
		$262144^1$&
			2.3$\times$$10^{-5}$&
					389.26&
							1.00\\
		&
		&
		$256/8$&
		$1024^8$&
			1.1--2.8&
					410.60&
						1.00
							\\
		\midrule
		\multirow{2}{*}{Hard-Gumbel PPL}
		&
		\multirow{2}{*}{This\,Work}
		&
		$32/1$&
		$1024^1$&
			\bftab 100.0&
					15.84&
						78.59\\
		&
		&
		$256/8$&
		$1024^8$&
			\bftab 100.0--100.0&
					2.17&
						193.53
							\\
	\midrule
		\bftab Softmax KNN-CE
		&
		\multirow{2}{*}{This\,Work}
		&
		$32/1$&
		$1024^1$&
			\bftab 100.0&
					13.46&
						88.59
							\\
		\multicolumn{1}{r}{($K/4 = 1$)}&
		&
		$256/8$&
		$1024^8$&
			\bftab 100.0--100.0&
					\bftab 1.57&
						\bftab 194.51
							\\
	\bottomrule
	\end{tabular}
	}
\end{table*}
\addtocounter{footnote}{-1}
\footnotetext{
	Attempts to replicate STE Euclid results using the official repository yielded highly degraded scores; for example, the FID scores were $172.78$ for $M=1024$ and $349.18$ for $M=16384$.
	These results are consistent with community-reported issues (e.g., \url{https://github.com/CompVis/taming-transformers/issues/87}).
	Furthermore, the pre-trained weights for the RE Euclid model \citep[for replicating Table~2 in][]{Fifty+25} are unavailable in the official repository (\url{https://github.com/cfifty/rotation_trick/tree/main}).
	Due to these reproducibility challenges, the ``optimal scores'' published in the original papers are reported for the single-codebook baselines.
}
\addtocounter{footnote}{1}
\footnotetext{
	The original SimVQ paper evaluated the method on $128 \times 128$ inputs while keeping a $16 \times 16$ latent map \citep[][]{Zhu+25}, which severely reduces the compression rate and makes reconstruction significantly easier. Since those scores are not directly comparable to the standard $256 \times 256$ VQGAN benchmark, SimVQ is evaluated here on the same $256 \times 256$ pipeline to ensure a rigorous and direct comparison.
}


Table~\ref{tab:score_vqgan} reports the codebook utilization and reconstruction performance for the proposed method (``KNN-CE'') and the baselines.
The reconstruction scores reported in the original studies are listed alongside those obtained using the corresponding methods in the present experimental setup.%

Just as in the discrete autoencoding, the proposed regularization successfully prevented code collapse, ensuring 100\% utilization of the codebook(s).
It also outperformed the existing methods when combined with the product quantization and equipped with a huge effective codebook size of $1024^8$.
Note that product quantization does not require a larger codebook matrix and the number of learnable parameters remains identical to the single codebook setting.

\section{Discussions}
\label{sec:discussions}


This study introduced a simple and unified regularization method that simultaneously tightens smoothed vector quantization and promotes effective code utilization.
The proposed method successfully prevented code collapse in both discrete autoencoding (\textsection\ref{sec:ae}) and VQGAN (\textsection\ref{sec:vqgan_image}).
This robustness is noteworthy, as many prior approaches were effective only in specific settings and remained vulnerable to code collapse in others.

The proposed method is geometrically intuitive and straightforward, yet appears to be unaddressed in the existing literature.
Research on neural vector quantization has traditionally been rooted in variational autoencoding \citep{KingmaWelling14_VAE}, with the primary aim of extending this stochastic framework to discrete variables \citep{Jang+17,vandenOord+17_VQVAE}.
The issue of code collapse was recognized (or documented) later, and workarounds were developed independently of the quantization methods themselves \citep{Dieleman+18,Baevski+20_wav2vec2.0,Dhariwal+20_Jukebox}.

By contrast, the present work reformulates neural vector quantization as a simple smoothing problem: onehot vectors are approximated by elements of the simplex.
Within this perspective, concentrating the approximators near the simplex vertices naturally arises as a desirable property.
This reformulation, together with the proposed regularization strategy, represents a key conceptual contribution of the study.

The KNN-based regularization can also overcome its limitation in the available codebook size by utilizing product quantization (\textsection\ref{sec:vqgan_image}).
While the maximum size of a single codebook is bounded by that of the batch and latent feature map, combining multiple codebooks can enlarge the effective vocabulary without requiring extra learnable parameters.
This workaround broadens the applicability of the proposed method.

\section*{Acknowledgments}
\lccode`\0`\0
\lccode`\1`\1
\lccode`\2`\2
\lccode`\3`\3
\lccode`\4`\4
\lccode`\5`\5
\lccode`\6`\6
\lccode`\7`\7
\lccode`\8`\8
\lccode`\9`\9
\hyphenation{JP-24-H-0-0-7-7-4 JP-22-H-0-3-9-1-4 JP-24-K-1-5-0-8-7 JP-MJCR-25-U6 JP-MJCR-22-P5 K35-XXVIII-620}
This study was supported by 
JSPS Grant-in-Aid for Scientific Research
A (JP24H00774, JP26H01985),
B (JP22H03914),
and C (JP24K15087);
JST AIP Accelerated Program (JPMJCR25U6) and Core Research for Evolutional Science and Technology (JPMJCR22P5);
Kayamori Foundation of Informational Science Advancement (K35XXVIII620);
and Kawai Foundation for Sound Technology \& Music.
The author also gratefully acknowledges the support of the
Academic Center for Computing and Media Studies, Kyoto University,
regarding the use of their supercomputer system.


\bibliographystyle{apalike}
\bibliography{takashi_references}

\appendix
\section{Implementation of the Quantization Methods}
\label{sec:implement_detail}

This section provides details on the implementation of the quantization methods.

\subsection{Smoothed Quantization}
\label{sec:smooth_detail}

Smoothed code-selection vectors $\mathbf{p}$ were computed as $\mathbf{p} = \operatorname{softmax}(\hat{Q}^{\mathsf{T}}\hat{\mathbf{z}}/\mathfrak{t})$.
In other words, the codebook vectors $(\mathbf{q}_1,\dots,\mathbf{q}_M) = Q$ and the feature vectors $\mathbf{z}$ were first L2-normalized, and their product (i.e., cosine similarity) was scaled by a learnable temperature $\mathfrak{t}$.
This temperature was shared across the codebook so that all the logits had the same amplitude.
Assignment probabilities $\mathcal{\pi}$ for Gumbel-softmax sampling were computed in the same way,
while the additional temperature parameter---$\tau$ in Equation~\ref{eq:gumbel-softmax}---was fixed as 1.0.%
\footnote{
	Previous studies manually annealed the Gumbel-softmax temperature from $\tau=0.5$ to $2.0$, scaling it by $0.999995$ at each iteration \citep[][]{Baevski+20_wav2vec2.0}.
	This approach was also tested in the experiments here but did not yield improvements over the fixed temperature.
}

\subsection{Hard Quantization}

The weight $\beta$ on the commitment loss in Equation~\ref{eq:vq-loss} was set to 1.0 \citep{Fifty+25,Zhu+25}, based on previous observations that its value does not significantly affect learning outcomes within the range 0.1--2.0. \citep{vandenOord+17_VQVAE}.

\section{Details of the Experiments}
\label{sec:experiment_detail}

\subsection{Discrete Autoencoding}
\label{sec:ae_detail}

This section provides implementation details for the autoencoding experiment described in \textsection\ref{sec:ae}.

The network architecture followed prior work on discrete autoencoding of ImageNet \citep{Esser+21,Fifty+25}.
Input images were center-cropped to $H \times W \times C = 256 \times 256 \times 3$.
The encoder first expanded the channel dimensionality of the input images from 3 to 256 by convolution, and then progressively downsampled them through a series of strided convolutions (see Table~\ref{tab:param_ae} for the spatial and channel sizes at each layer).
Each downsampling layer was followed by a residual block \citep{He+16_ResNet}.
The decoder reconstructed the input images by upsampling the latent feature maps with a sequence of interpolations and residual blocks.

To improve memory efficiency,
all but the input and output layers were implemented as depthwise separable convolutions \citep{Chollet17_Xception}.
All convolutional kernels had size $3 \times 3$.

Training employed the AdamW optimizer with $(\beta_1,\beta_2)=(0.9,0.99)$ and
a weight decay coefficient of $10^{-4}$,
except for Euclidean-based STE/rotational hard quantization, where weight decay was set to zero.

The learning rate was linearly warmed up from $0.0$ to $\rho_{\text{max}}$, and subsequently annealed to $0.5\rho_{\text{max}}$ by cosine scheduling.
For discrete autoencoding, the maximum learning rate $\rho_{\text{max}}$ was set to $5 \times 10^{-5}$ for Euclidean-based STE/rotational hard quantization, and to $10^{-4}$ for all other configurations \citep{Fifty+25}.

\begin{table*}
	\centering
	\caption{Hyperparameters for discrete autoencoding/GAN.}
	\label{tab:param_ae}
	{\fontsize{7.5}{8.5}\selectfont
	\begin{tabular}{lrrrr}
	\toprule
	&
		GAN (\textsection\ref{sec:vqgan_image})&
			\multicolumn{3}{c}{Autoencoding (\textsection\ref{sec:ae})}
			\\
	\cmidrule(l{0.4em}r{0.4em}){2-2}
	\cmidrule(l{0.4em}r{0.4em}){3-5}
	Feature Map Size&
		$16 \times 16 \times 32$
		&
		$16 \times 16 \times 32$
		&
			$64 \times 64 \times 3$
			&
			$64 \times 64 \times 64$
				\\
	Codebook Size&
		1024&
			8196&
			8196&
				8196
				\\
	Latent Channels&
		\multicolumn{2}{c}{
		$128 \to 128 \to 64 \to 64 \to 32 \to 32$
		}
		&
			\multicolumn{2}{r}{
			$128 \to 64 \to 32 \to \{3,64\}$
			}
				\\
	Height \& Width&
		\multicolumn{2}{c}{
		$256 \to 128 \to 64 \to 32 \to 16 \to 16$
		}
		&
			\multicolumn{2}{r}{
			$256 \to 128 \to 64 \to 64$
			}
				\\
	\midrule
	Batch Size&
		64&
		64&
			64&
				64\\
	Training Epochs&
		60&
		25&
			20&
				20\\
	Warmup Iterations&
		20,000&
		16,000&
			16,000&
				16,000\\
	\bottomrule
	\end{tabular}
	}
\end{table*}

\subsection{VQGAN}
\label{sec:vqgan_detail}

In the VQGAN experiment (\textsection\ref{sec:vqgan_image}), the models were optimized based on the following learning objective:

\begin{align}
	\mathcal{L}_{\text{main}}
		=
			\underbracket{
				| x - \mathfrak{F}(x) |_1
			}_{\text{Reconstruction}}
			+
			\overbracket{
			\sum_{l} \| \mathfrak{P}_l(x) - \mathfrak{P}_l\circ \mathfrak{F}(x) \|_2^2
			}^{\text{Perceptual}}
			\underbracket{
			-
			\lambda
			\mathfrak{G} \circ \mathfrak{F}(x)
			}_{\text{Adversarial}}
\end{align}
$\mathfrak{F}$ denotes the model network reconstructing original images $x$ through discrete latent representations.
$\mathfrak{P}$ is a perceptual-loss function implemented by the VGG-16 network pretrained on image classification and finetuned to predict image similarity judgements by human evaluators \citep{Johnson+16,Zhang+18_PerceptualLoss}.
$\mathfrak{G}$ is an adversarial discriminator network \citep[PatchGAN implementation;][]{Isola+17}, which was trained with the hinge loss Equation~\ref{eq:hinge}.

\begin{align}
	\mathcal{L}_{\text{disc}}
		:=
			\max\{0, 1 - \mathfrak{G}(x)\}
			+
			\max\{0, 1 + \mathfrak{G}\circ\mathfrak{F}(x)\}
			\label{eq:hinge}
\end{align}

The weight $\lambda$ on the adversarial loss was set to zero during the initial 600k iterations, and to 0.1 for the rest of training.
The maximum learning rate $\rho_{\text{max}}$ for the main network was set to $10^{-4}$,
whereas that for the discriminator was set to $4.5 \times 10^{-6}$.

\subsection{Evaluation Metrics}

Fr\'{e}chet Inception Distance \citep[FID;][]{Heusel+17} and Inception Score \citep[IS;][]{Salimans+16} were evaluated using PyTorch Ignite implementations \citep{pytorch-ignite}.
The weights of their feature extractor network (Inception V3) conformed to the original implementation of FID.
%

\begin{table}
	\caption{Wall-clock time and peak VRAM consumption per training iteration. The models were run on four NVIDIA A100 GPUs (80GB VRAM per GPU) using \texttt{DistributedDataParallel} of PyTorch. The reported scores represent computational load per GPU on average over a single epoch of the ImageNet training split, with batch size set to 64. The ``Entire'' wall-clock time includes the forward and backward passes through the encoder, quantizer, and decoder modules, while the `Quant.' reports the wall-clock time consumed by the forward pass through the quantizer alone.
	}
	\label{tab:time-and-memory}
	\centering
	{\fontsize{6.5}{7.5}\selectfont
	\begin{tabular}{llcrrrrrrrrr}
	\toprule
	&&&
		\multicolumn{9}{c}{Feature\,Map\,Size; Codebook\,Size}
		\\
	&&&
		\multicolumn{3}{c}{16$\times$16$\times$32; 1024}
		&
		\multicolumn{3}{c}{64$\times$64$\times$3; 8196}
		&
		\multicolumn{3}{c}{64$\times$64$\times$64; 8196}
		\\
		\cmidrule(l{0.4em}r{0.4em}){4-6}
		\cmidrule(l{0.4em}r{0.4em}){7-9}
		\cmidrule(l{0.4em}r{0.4em}){10-12}
	&&&
		\multicolumn{2}{c}{Wall-Clock Time}
		&
			&
		\multicolumn{2}{c}{Wall-Clock Time}
		&
			&
		\multicolumn{2}{c}{Wall-Clock Time}
		&
			\\
		\cmidrule(l{0.4em}r{0.4em}){4-5}
		\cmidrule(l{0.4em}r{0.4em}){7-8}
		\cmidrule(l{0.4em}r{0.4em}){10-11}
	\multicolumn{2}{l}{Method}&
	$K/4$&
		Entire&
			Quant.&
				VRAM&
		Entire&
			Quant.&
				VRAM&
		Entire&
			Quant.&
				VRAM\\
	&&&
		(ms)&
			(ms)&
				(GB)&
		(ms)&
			(ms)&
				(GB)&
		(ms)&
			(ms)&
				(GB)\\
	\midrule
	\multirow{2}{*}{\rotatebox{90}{STE}}
	&
	STE Euclid
	&
	---&
		330.39&
			1.35&
				34.20&
		508.49&
			1.35&
				54.83&
		502.20&
			1.30&
				54.93\\
	&
	SimVQ
	&
	---&
		330.18&
			1.31&
				34.20&
		504.28&
			1.36&
				54.83&
		502.25&
			1.37&
				54.92\\
	\midrule
	\multicolumn{2}{l}{Hard-Gumbel PPL}
	&
	---&
		324.95&
			1.94&
				34.24&
		615.49&
			1.89&
				59.83&
		614.21&
			1.92&
				59.89\\
	\midrule
	\multirow{4}{*}{\rotatebox{90}{Softmax}}
	&
	No Regularization
	&
	---
	&
		323.48&
			1.40&
				34.24&
		561.05&
			1.44&
				59.83&
		557.22&
			1.38&
				59.87\\
	\cmidrule{2-12}
	&
	PPL
	&
	---
	&
		324.99&
			1.68&
				34.24&
		601.19&
			1.64&
				59.83&
		598.18&
			1.57&
				59.87\\
	\cmidrule{2-12}
	&
	\multirow{2}{*}{KNN-CE}
	&
	1
	&
		324.92&
			1.55&
				34.24&
		596.11&
			1.58&
				59.83&
		592.39&
			1.58&
				59.87\\
	&&
	8
	&
		325.60&
			1.50&
				34.24&
		595.86&
			1.64&
				59.83&
		592.85&
			1.54&
				59.87\\
	\bottomrule
	\end{tabular}
	}
\end{table}

\section{Computational Load}
\label{sec:compute-load}

To evaluate the computational load of the proposed method, the wall-clock time and peak VRAM consumption per training iteration were measured during the discrete autoencoding task (Table~\ref{tab:time-and-memory}).
The models were run on four NVIDIA A100 GPUs (80GB VRAM per GPU) using \texttt{DistributedDataParallel} of PyTorch. The reported scores represent computational load per GPU on average over a single epoch of the ImageNet training split, with batch size set to 64.

This benchmark results suggest that the proposed regularization method (``KNN-CE'') is not computationally expensive compared to existing methods in the (Gumbel-)Softmax family.
The execution time is almost identical to the perplexity-based regularization (``PPL''),
and is smaller than its combination with the hard Gumbel-Softmax sampling.
Similarly, the VRAM consumption is identical across the smoothing approaches, indicating that the major source of memory load is in the softmax computation, not in the regularization.

\section{Alternative Implementations of the Intended Regularization}
\label{sec:dirichlet}

\begin{figure}
	\centering
	\includegraphics[width=\columnwidth]{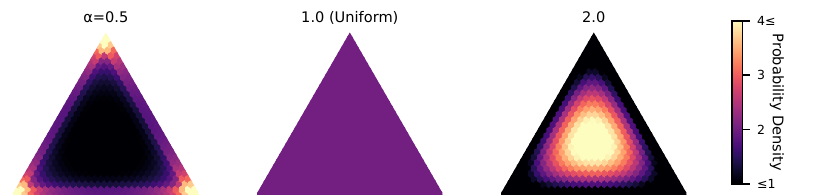}
	\caption{Dirichlet distributions on the simplex $\Delta^{3-1}$ with concentration parameters $\alpha_1=\alpha_2=\alpha_3=\alpha$, where $\alpha \in \{0.5, 1.0, 2.0\}$.
			}
	\label{fig:dirichlets}
\end{figure}

Alternative regularization strategies could also achieve the intended distribution of smoothed code-selection vectors ($\mathbf{p}$ or $\boldsymbol{\pi}$) around the simplex vertices.
For example, one could align smoothed code-selectors with a Dirichlet distribution with concentration parameters $\alpha_1=\cdots=\alpha_K<1.0$ (Figure~\ref{fig:dirichlets}).
The probability density of such a distribution is highest at the vertices of the simplex, thus matching the ideal distribution of smoothed code-selectors.
A possible formalization of this alignment is based on the Kullback-Leibler (KL) divergence between the Dirichlet prior ($\mathbb{P}$) and the distribution inferred from smoothed code-selectors ($\mathbb{Q}$).
\begin{align}
	\mathcal{D}_{\text{KL}} \left(\mathbb{P} \mid \mathbb{Q}\right)
		:=&
			\int_{\Delta^{M-1}} \mathbb{P}(\mathbf{p}) \log \frac{\mathbb{P}(\mathbf{p})}{\mathbb{Q}(\mathbf{p})} d\mathbf{p}
			\label{eq:KL}
\end{align}

However, Equation~\ref{eq:KL} is difficult to use directly as a regularization loss, due to the complexity of estimating $\mathbb{Q}$ from sample code-selectors, $\mathbf{p}$.
Although one could constrain $\mathbb{Q}$ as another Dirichlet to make the KL divergence tractable,
maximum likelihood estimation of its parameters requires iterative algorithms \citep[e.g., the Newton-Raphson method;][]{Ronning89,Sklar14,Wicker+08}, complicating and slowing gradient-based optimization in deep learning frameworks.
Moreover, this estimation involves digamma and trigamma functions \citep{Sklar14}, whose derivatives can explode when the concentration parameters approach small values---as desired for tight smoothing---during the course of learning.

A more practical approach is to approximate Equation~\ref{eq:KL} itself in a tractable manner.
For instance, \citet{Perez-Cruz08} proposed a KNN-based estimation of the KL divergence that relies solely on \emph{samples} from the two distributions, using the $k$-th nearest neighbor from $\mathbb{Q}$ for each sample from $\mathbb{P}$.
The regularization method proposed in this study can thus be interpreted as minimizing this estimated KL divergence, with the samples from $\mathbb{P}$ constrained to onehot vectors.

\end{document}